\documentclass[letterpaper, 10 pt, conference]{ieeeconf}  

\IEEEoverridecommandlockouts                              

\usepackage{graphics} 
\usepackage{epsfig} 
\usepackage{times} 
\usepackage{amsmath} 
\usepackage{amssymb}  
\usepackage{caption}
\usepackage{lipsum}
\usepackage{float}

\usepackage{algorithm}
\usepackage{algorithmic}
\usepackage[table]{xcolor}
\usepackage{xcolor}         
\usepackage{booktabs}
\usepackage{subfigure}
\usepackage{setspace}
\usepackage{bm}
\usepackage{color}
\usepackage{multirow}
\usepackage{lipsum}

\definecolor{myblue}{rgb}{0.854, 0.890, 0.953} 
\definecolor{mygreen}{rgb}{0.886, 0.941, 0.851} 
\definecolor{myred}{rgb}{0.984, 0.898, 0.839}

\title{\LARGE \bf
Improving Vision-Language-Action Model with \\ Online Reinforcement Learning
}

\author{Yanjiang Guo$^{13*}$, Jianke Zhang$^{1*}$, Xiaoyu Chen$^{13*}$, Xiang Ji$^{1}$, Yen-Jen Wang$^{2}$, Yucheng Hu$^{1}$, Jianyu Chen$^{13\dagger}$
\thanks{*This work was not supported by any organization}
\thanks{$^{1}$Albert Author is with Faculty of Electrical Engineering, Mathematics and Computer Science,
        University of Twente, 7500 AE Enschede, The Netherlands
        {\tt\small albert.author@papercept.net}}%
\thanks{$^{2}$Bernard D. Researcheris with the Department of Electrical Engineering, Wright State University,
        Dayton, OH 45435, USA
        {\tt\small b.d.researcher@ieee.org}}%
}

\begin{document}

\twocolumn[{%
\renewcommand\twocolumn[1][]{#1}%
\maketitle
\begin{center}
    \centering
    \vspace*{-5mm} 
    \includegraphics[width=1.00\linewidth]{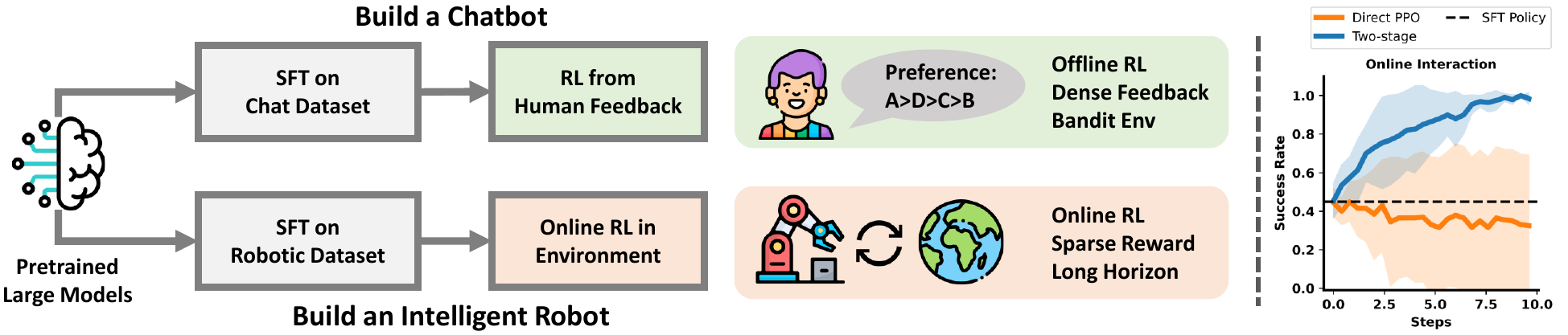}
    \vspace*{-3mm}
    \captionof{figure}{Illustration of our motivation. We employ the fine-tuning pipeline from large language models (LLMs) to enhance the Vision-Language Architecture (VLA) in the robotic domain, starting with supervised fine-tuning (SFT) followed by reinforcement learning (RL).
    However, we observed that standard online RL can be extremely unstable when applied to large VLA models. To address this, we propose an iterative RL method, iRe-VLA.}
    \label{motivation}
\end{center}%
}]
\thispagestyle{empty}
\pagestyle{empty}

\begin{abstract}

Recent studies have successfully integrated large vision-language models (VLMs) into low-level robotic control by supervised fine-tuning (SFT) with expert robotic datasets, resulting in what we term vision-language-action (VLA) models. Although the VLA models are powerful, how to improve these large models during interaction with environments remains an open question. 
In this paper, we explore how to further improve these VLA models via Reinforcement Learning (RL), a commonly used fine-tuning technique for large models. 
However, we find that directly applying online RL to large VLA models presents significant challenges, including training instability that severely impacts the performance of large models, and computing burdens that exceed the capabilities of most local machines. To address these challenges, we propose iRe-VLA framework, which iterates between Reinforcement Learning and Supervised Learning to effectively improve VLA models, leveraging the exploratory benefits of RL while maintaining the stability of supervised learning. Experiments in two simulated benchmarks and a real-world manipulation suite validate the effectiveness of our method. 

\end{abstract}

{
\renewcommand{\thefootnote}
{\fnsymbol{footnote}}
\footnotetext{$^*$Equal contribution}

\renewcommand{\thefootnote}
{\fnsymbol{footnote}}
\footnotetext{$^\dagger$Corresponding author.{\tt\footnotesize jianyuchen@tsinghua.edu.cn}}

\renewcommand{\thefootnote}1
{\fnsymbol{footnote}}
\footnotetext[1]{Institute for Interdisciplinary Information Sciences, Tsinghua University, Beijing, China. {\tt\footnotesize guoyj22@mails.tsinghua.edu.cn}}

\renewcommand{\thefootnote}2
\footnotetext[2]{University of California, Berkeley, USA.}

\renewcommand{\thefootnote}3
\footnotetext[2]{Shanghai Qi Zhi Institute, Shanghai, China.}
}
\section{Introduction}
It has become a recent trend to employ powerful pre-trained large language models (LLMs) and vision-language models (VLMs) for a variety of advanced tasks beyond their original scope, including dialogue systems \cite{ouyang2022training,glaese2022improving,thoppilan2022lamda}, code generation \cite{chen2021evaluating}, task planning \cite{ahn2022can,huang2022inner}, and even low-level robotic control \cite{brohan2022rt,brohan2023rt}. By fine-tuning VLMs on robotic datasets with explicit action modeling, previous works have developed large vision-language-action (VLA) models \cite{ma2024survey}, such as RT-2 \cite{brohan2023rt}, HiRT\cite{zhanghirt}, Roboflamingo \cite{li2023vision}, etc. These models are capable of directly outputting low-level robotic control signals while also benefiting from the common-sense knowledge and reasoning abilities \cite{wei2022chain} encoded in large pre-trained models.

The fine-tuning of VLA models generally employs a supervised fine-tuning (SFT) approach \cite{brohan2023rt}, noted for its stability and scalability. However, SFT depends on high-quality expert datasets that are costly and difficult to obtain in the robotic domain \cite{padalkar2023open}. Additionally, supervised learning may not fully align VLA models with physical environments due to distribution shift issues \cite{belkhale2024data,kumar2020conservative}. 
We wonder how to further improve such large VLA models through interaction with the physical environment beyond supervised learning. 
Notably, Reinforcement Learning from Human Feedback (RLHF) \cite{ouyang2022training,liu2020learning,christiano2017deep} has better align large language model with human preference, as illustrated in the upper-left of Figure \ref{motivation}. 

Inspired by the success of RLHF, we try online RL to improve the VLA model and better align the VLA model with physical environments. However, the environments encountered by chatbots and embodied robots are markedly different. Chatbots are optimized using offline, human-labeled datasets with well-defined dynamics \cite{ouyang2022training}, while embodied robots necessitate online exploration in tasks characterized by long horizons and sparse rewards. 
Furthermore, previous research has shown that the \textbf{online reinforcement learning (RL) process can be extremely unstable when applied to large neural networks}\cite{parisotto2020stabilizing,andrychowicz2020matters,ota2021training}. 
Empirically, we also observe that directly applying the standard RL algorithm to large VLA models results in training instability and performance drops, as depicted on right side of Figure \ref{motivation}. 
To stabilize the RL process and effectively enhance the VLA model, we propose the novel \textbf{iRe-VLA} method, which \textbf{i}terates between online \textbf{Re}inforcement Learning stages and supervised learning stages. 
Specifically, during the RL stage, we freeze the VLM parameters and only train lightweight action heads to maintain training stability. In the subsequent supervised learning phase, we fine-tune the entire model on successful trajectories to fully utilize the expressive capabilities of the large model. Empirically, this two-stage approach consistently enhances the VLA's performance, stabilizes training, and is computationally more efficient. We have validated the iRe-VLA methods through comprehensive experiments, including simulated MetaWorld \cite{yu2020meta}, Franka-Kitchen \cite{gupta2019relay}, and real-world Panda manipulation task sets. In these domains, our method not only better aligns the VLA model with the original tasks but also autonomously solves unseen tasks. Furthermore, the VLA model's generalization ability has also been improved through online interactions with the environment.

\section{Related Works}

\textbf{Foundation Models for Embodied control. }
Large Language Models (LLMs) and vision-language models (VLMs) trained on web-scale data encode knowledge of the physical world and exhibit impressive reasoning ability. With this prior knowledge, LLMs and VLMs can benefit the embodied control tasks in many aspects, ranging from providing rewards or values \cite{ma2023eureka,fan2022minedojo,adeniji2023language} for agents, modeling the world dynamics \cite{lin2023learning,hanjie2021grounding}, or directly as policy \cite{ahn2022can,huang2022inner,zeng2022socratic,driess2023palm,guo2023doremi,dasgupta2023collaborating,wang2023prompt}. 

As for literature using LLMs/VLMs directly as agents' policy, we can roughly divide them into two categories, namely high-level planning and low-level control. Works in the first categories leverage LLMs' reasoning ability to autoregressively generate the textual step sequences \cite{ahn2022can,huang2022inner,zeng2022socratic} or code \cite{liang2022code}, thereby decomposing the long-horizon tasks into feasible plans. However, these methods output textual plans that are not directly grounded in the physical world and require powerful low-level skills. Another line of work leveraged VLMs to directly output low-level control signals and verified that low-level skills themselves could also benefit from the prior knowledge encoded in the pre-trained VLMs \cite{brohan2022rt,brohan2023rt,zhanghirt,chen2024vision,li2023vision}. Since the original output of VLMs lies in the language space, these works need additional action modeling parts like adding action heads \cite{zhanghirt,li2023vision} or replacing the language tokens with actions \cite{brohan2023rt}.

\textbf{Finetune Large Models with RL. }
Reinforcement learning has been successfully used in the natural language process downstream tasks to better align the generated text to human preferences \cite{ouyang2022training,stiennon2020learning,ramamurthy2022reinforcement}. In this Reinforcement Learning from Human Feedback (RLHF) framework, a reward model is trained on a pre-collected human preference dataset and then LLM is optimized in a bandit environment with constraints of not shifting too much from the original model \cite{ouyang2022training}, which can be seen as offline-style RL \cite{levine2020offline}. Different from RLHF for dialog systems, fine-tuning VLA models face unknown dynamics and require online exploration \cite{carta2023grounding,szot2023large,zhai2024fine}. For instance, GLAM \cite{carta2023grounding} ground the LLM textual plans in simplified grid-world environments through online RL. LLaRP \cite{szot2023large} ground the high-level plans generated by VLMs in rearrangement tasks with dense reward RL. However, they all assume low-level skills (e.g., pick, goto) are available and only better ground the high-level plans. Different from them, we try to use RL to directly improve the low-level control signal output by VLA policy which has much longer horizons (hundreds or thousands of steps) in sparse-reward physical environments.

\begin{figure*}[t]
    \centering
    \includegraphics[width=1.0\textwidth]{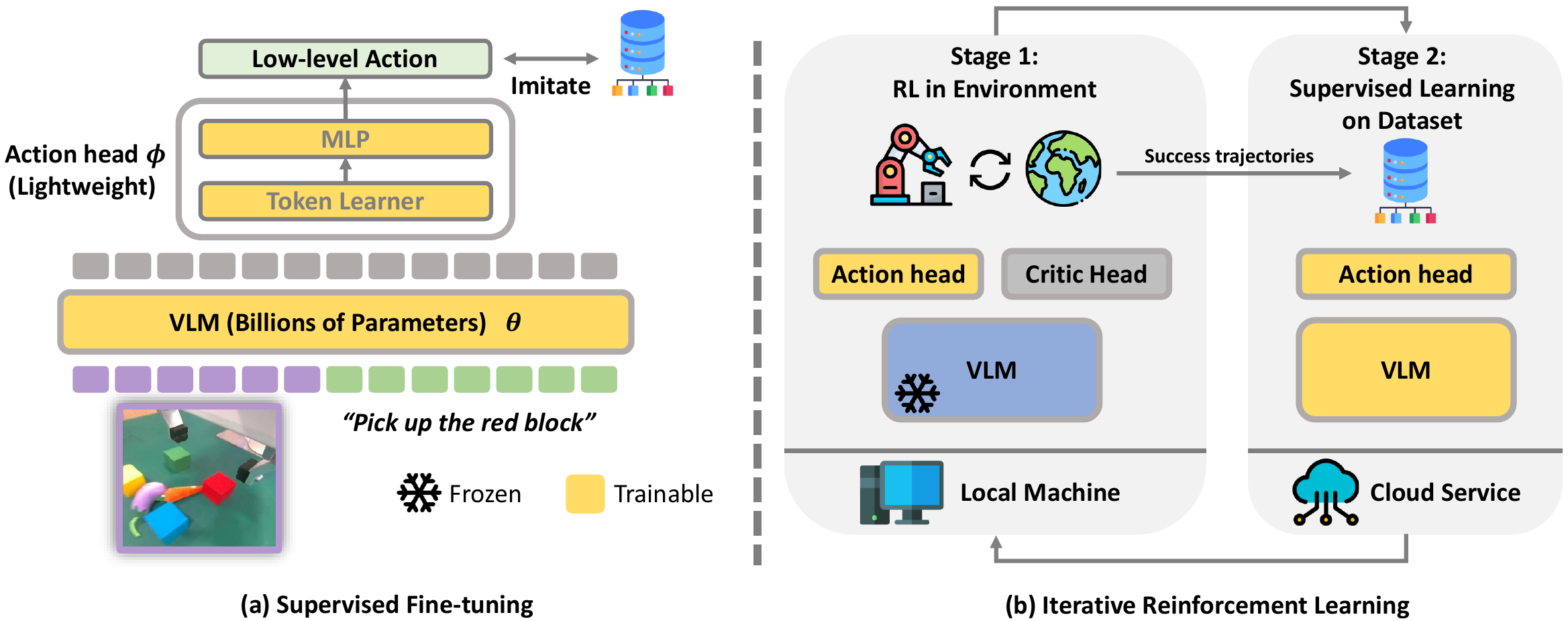}
    \caption{(a) Our VLA model comprised a pre-trained VLM backbone and lightweight action head. (b) During the finetuning, we iterate between exploration and SL stages to effectively improve the VLA model. The VLM is frozen in the exploration stage to stabilize training and trainable in the SL stage to fully leverage the power of pre-trained VLM.}
    \label{method}
    \vspace{-1mm}
\end{figure*}
\section{Preliminary}

\textbf{Reinforcement Learning. }
We utilize the standard deep RL partially-observed Markov decision process (POMDP) framework, where a task can be modeled as $\mathcal{M}=(\mathcal{S},\mathcal{A},P_{T},R,\gamma,\mathcal{O},P_{E})$. $\mathcal{S}$ and $\mathcal{A}$ are the state space and action space for tasks, $\mathcal{O}$ is the robot observation, such as visual image. $P_T:\mathcal{S}\times \mathcal{A}\times \mathcal{S} \rightarrow [0,1]$ are state transition probability functions and $R:\mathcal{S}\times \mathcal{A}\times \mathcal{S} \rightarrow \mathbb{R}$ are reward function for the task. In robotic tasks, the reward signal is always sparse, so we consider binary reward in this paper, where $R=1$ if the robot successfully finished the task otherwise $R=0$. $P_E: \mathcal{S}\times \mathcal{O} \rightarrow [0,1]$ is the observation emission probabilities. A policy $\pi_{\theta}: \mathcal{O}\rightarrow \mathcal{A}$ defines a probability distribution in action space parameterized by $\theta$. The objective of parameter $\theta$ is to maximize the expected return of the policy $\pi_{\theta}$ with discount $\gamma$:
\begin{equation}
    J(\theta)=\mathbb{E}_{((s_0,o_0,a_0),(s_1,o_1,a_1),...)\sim p_{\theta}} \left[ \sum_t \gamma^t R(s^t,a^t) \right]
\end{equation} 

\textbf{Vision-Language Model. }
Numerous vision-language models (VLMs) have been developed that can concurrently process visual and language input. These models can broadly be classified into two categories \cite{brohan2023rt}: representation learning models, such as CLIP \cite{radford2021learning}, and generative models, such as Blip-2 \cite{li2023blip} and InstructBlip \cite{dai2024instructblip}.
Following \cite{brohan2023rt,chen2024vision,li2023vision}, we particularly employ the generative VLMs in the format of \{vision, text\}$\rightarrow$\{text\}. Formally, the generative VLMs sample tokens $x^{1:K}$ from $p(x^{1:K}|I,c)$, which are conditioned on the input image $I$ and instruction $c$. Since original generative VLMs produce natural language outputs, integrating these models into robotic control tasks requires an additional action modeling component, detailed in the subsequent section.

\section{Method}
Our goal is to develop a learning method that effectively improves the VLA model through online interactions while maintaining computational costs affordable for robotic systems. We start with a Vision-Language-Action (VLA) model fine-tuned on robotic demonstrations. We detail the VLA architectures in Section \ref{4.1} and outline the learning pipeline of the iRe-VLA method in Section \ref{4.2}.

\subsection{Model Architectures}\label{4.1}
Our VLA model transforms vision input $o \in \mathcal{O}$ and free-form language instruction $i \in \mathcal{L}$ into low-level robotic action $a \in \mathcal{A}$, represented as $\mathcal{O} \times \mathcal{L} \rightarrow \mathcal{A}$. The model comprises a pre-trained large VLM and a lightweight action head, as illustrated on the left side of Figure \ref{method}. 

We utilize the BLIP-2 3B model \cite{li2023blip} as our backbone VLM. Since pre-trained VLM output text tokens in language space, an action head is designed to produce low-level control actions. These actions typically include changes in the end-effector's pose and the gripper's status. Following the design presented in \cite{li2023vision,chen2024vision}, we replace the VLM's final fully connected layer with a newly initialized action head. In the action head, a token learner \cite{lee2019set} first converts the VLM’s last hidden representation $h \in \mathbb{R}^{m\times d}$ to $h' \in \mathbb{R}^{d}$. Subsequently, a Multi-Layer-Perceptron (MLP) \cite{riedmiller2014multi} map $h'$ to the action $a \in \mathbb{R}^{d_a}$, where $m$ and $d$ denote the number of tokens and the embedding dimension of the VLM, respectively, and $d_a$ represents the action dimensions.

\textbf{Low-Rank Adaptation(LoRA) \cite{hu2021lora}}
Our VLA model comprises a large VLM backbone and a lightweight action head. However, fine-tuning the entire model, with its billions of parameters, requires significant computational resources. Furthermore, previous studies \cite{liu2023tail,bousmalis2023robocat} suggest that fine-tuning the whole large pre-trained model in limited-data regimens can result in over-fitting. Following the approach described in \cite{liu2023tail}, we utilize the parameter-efficient LoRA method to fine-tune the VLM part. The total trainable parameters consist of the LoRA parameters $\theta$ and the action head parameters $\phi$. 



\begin{figure}[t]
\vspace{-2mm}
\begin{algorithm}[H]
   \caption{\textbf{Iterative RL for VLA model (iRe-VLA)} }
   \label{algo1}
   \textbf{Given:} A expert dataset $D_e$, a supevise fine-tuned VLA model $\pi^0_{\theta,\phi}$ with VLM parameters $\theta$ and action head $\phi$, \textbf{unseen tasks set $T=\{T_1,...T_n\}$}.
\begin{algorithmic}[1]
   \STATE{Initialize the online dataset $D_{RL} \leftarrow \emptyset$, copy the weight of $\pi^0_{\theta,\phi}$ to $\pi^1_{\theta,\phi},\pi^2_{\theta,\phi}$}
   \FOR {$T_i$ {\bfseries in} \{$T_0, T_1, ...,T_n$\}}
   \STATE{\textcolor[RGB]{100,100,220}{\#  Stage 1: RL}}
   \STATE{Copy the weight of $\pi^2_{\theta,\phi}$ to $\pi^1_{\theta,\phi}$, initialize a critic head.}
   \STATE{Optimize $\phi$ with online reinforcement learning until convergence by equation \ref{stage1}.}
   \STATE{Collect successful trajectories $x_i$ into $D_{RL}$:$D_{RL}=D_{RL}\cup x_i$.}
   \STATE{\textcolor[RGB]{100,100,220}{\#  Stage 2: SL}}
   \STATE{Copy the weight of $\pi^1_{\theta,\phi}$ to $\pi^2_{\theta,\phi}$.}
   \STATE{Optimize $\theta,\phi$ with supervised learning on $D_e \cup D_{RL}$ by equation \ref{stage2}.}
   \ENDFOR
\end{algorithmic}
\end{algorithm}
\vspace{-9mm}
\end{figure}

\subsection{Learning Pipeline} \label{4.2}
We desciribe the learning pipeline in this section. First, we supervised fine-tuning the VLA model on robotic datasets (stage 0), then we iterative between online RL (stage 1) and supervised learning (stage 2).

\textbf{Stage 0: Supervised Learning on Expert Dataset. }
We first perform standard supervised fine-tuning on the VLA model $\pi_{\theta}$ with the expert robotic dataset $D_e=\{(o_1,l_1,a_1),(o_2,l_2,a_2),...,(o_i,l_i,a_i)\}$. Formally, the learning objective is defined by a Mean Squared Error (MSE) loss:
\begin{equation}\label{stage0} 
    J^0(\theta,\phi)=\mathbb{E}_{(o,l,a)\sim D_e}\left[\left|\left|\pi_{\theta,\phi}(o,l)-a \right|\right|^2_2\right]
\end{equation}
After supervised fine-tuning, we obtain the initial VLA model $\pi^0_{\theta,\phi}$. The performance of $\pi^0_{\theta,\phi}$ is highly correlated to the scale and quality of the expert dataset $D_e$. Then we start to improve the $\pi^0_{\theta,\phi}$ through online RL.

\textbf{Stage 1: Online RL with Frozen VLM. } 
The SFT model, $\pi^0_{\theta,\phi}$, may not achieve optimal performance for new tasks. However, it serves as a valuable starting point since it has been trained on a variety of tasks from the robotic dataset. To enhance the performance of the SFT policy, we utilize online reinforcement learning (RL). In the RL process, we introduce a critic head that mirrors the structure of the action head, but with the output dimension set to one. To prevent model collapse and accelerate the learning process, we freeze the VLM parameters, $\theta$, during this phase. Consequently, only the parameters of the action head, $\phi$, are optimized:
\begin{equation}\label{stage1}
    J^1(\phi)=\mathbb{E}_{((s_0,o_0,a_0),(s_1,o_1,a_1),...)\sim p_{\phi}} \left[ \sum_t \gamma^t R(o^t,a^t) \right]
\end{equation} 
After online RL, the robot may discover new trajectories $x_i$ to solve new tasks. Then we collected these success trajectories into an online dataset $D_{RL}=D_{RL} \cup x_i$

\textbf{Stage 2: Supervised Learning on Both Expert and Online-collected Data. }
In Stage 1, while the agent conducts RL on new tasks, it risks forgetting previously learned tasks. Hence, in Stage 2, we supervise the whole model using both the newly collected online data $D_{RL}$ and the original expert dataset $D_e$ to mitigate catastrophic forgetting \cite{mccloskey1989catastrophic}. Formally, the objective can be written as:
\begin{equation}\label{stage2}
    J^2(\theta,\phi)=\mathbb{E}_{(o,l,a)\sim D_e \cup D_{RL}}\left[\left|\left|\pi_{\theta,\phi}(o,l)-a \right|\right|^2_2\right]
\end{equation} 
\textbf{Iterate between Stage 1 and Stage 2. }As previously noted, the agent in Stage 1 explores novel solutions for new tasks, while in Stage 2, it imitates all available success trajectories. By alternating between Stages 1 and 2, large VLA models progressively address a broader range of tasks while also preventing catastrophic forgetting on seen tasks. Furthermore, as suggested in previous works \cite{team2023octo,padalkar2023open}, the VLA model could become more generalizable by imitating a wider range of tasks. 
The whole pipeline is outlined in Algorithm \ref{algo1}.

\section{Experiments}
In this section, we perform tense experiments in two simulated benchmarks Metaworld and FrankaKitchen, and real-world panda manipulation tasks to verify the effectiveness of our iRe-VLA framework. We aim to answer the following questions: 
\begin{itemize}
    \item Why do we adopt a two-stage iterative RL process instead of standard RL?
    \item Can iRe-VLA stabilize the training process and effectively improve the VLA model in both expert tasks and unseen tasks?
    \item Can iRe-VLA lead to better generalization of the VLA model? 
\end{itemize}

\subsection{Experiment Setups}
\begin{figure*}[t]
    \centering
    \includegraphics[width=1.0\textwidth]{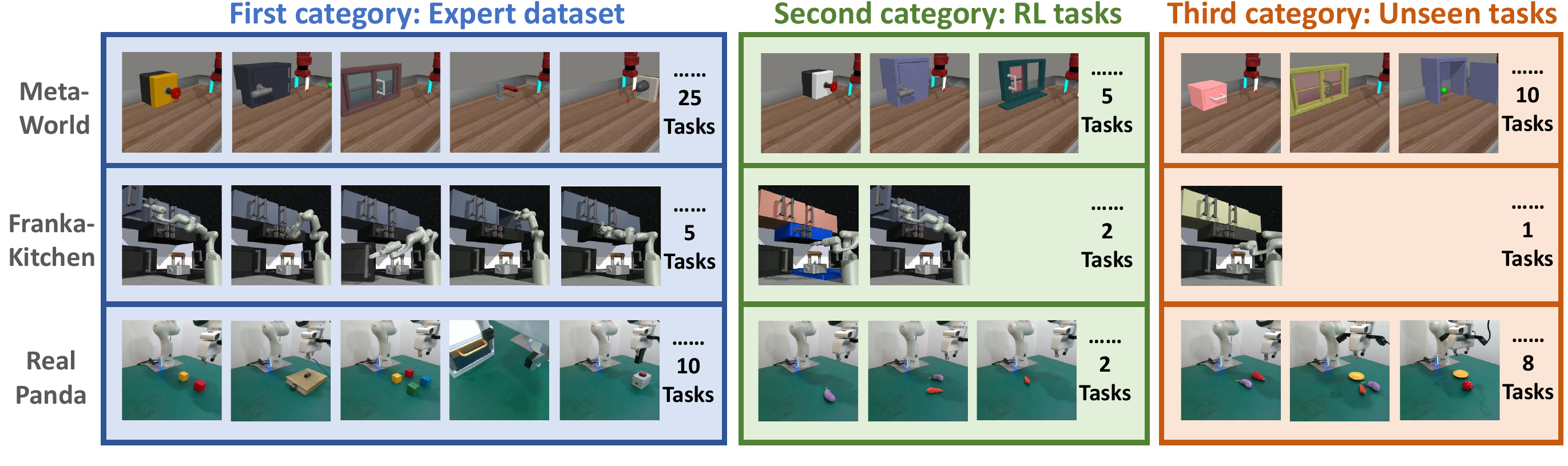}
    \caption{We perform experiments in three domains. Each domain encompasses three categories: tasks observed in the expert dataset, new tasks utilizing reinforcement learning, and hold-out unseen tasks. The tasks vary by required skills, as well as the shapes and appearances of objects. The initial positions of objects in each task are randomized in every episode.}
    \label{exp}
    \vspace{-4mm}
\end{figure*}

We perform experiments in three domains: Meatworld \cite{yu2020meta}, Franka Kitchen \cite{gupta2019relay}, and real-world panda manipulation, as illustrated in Figure \ref{exp}. Notably, we use \textbf{a single text-conditioned VLA model} to solve all tasks in a domain. 
Each domain involves tasks categorized into three groups: expert tasks observed in the demonstration datasets, RL-trained tasks enhanced by online RL, and hold-out tasks that are unseen in prior training. 
Initially, we conducted supervised fine-tuning on the VLA model using expert datasets. Subsequently, we improve the performance of the VLA model in second-category new tasks through online RL. Lastly, the third-category tasks are employed to evaluate the generalization capabilities of the trained VLA policy.

In the Metaworld domain, the expert dataset contains 25 tasks each with 50 trajectories. The second and third category introduces novel tasks featuring variations in object shape, color, and position. In the Franka kitchen domain, we follow the setting in \cite{liu2023tail}, the expert dataset contains 5 tasks while the tasks in the second and third categories encompass unseen changes in object appearance and position. As for real-world tasks, we collect 2,000 trajectories through teleoperation and script for picking (grasp), placing, button-press, cable-route, and drawer-open. The unseen tasks of real-world experiments include picking up unseen objects.




\begin{figure*}[t]
    \centering
    \includegraphics[width=1.0\textwidth]{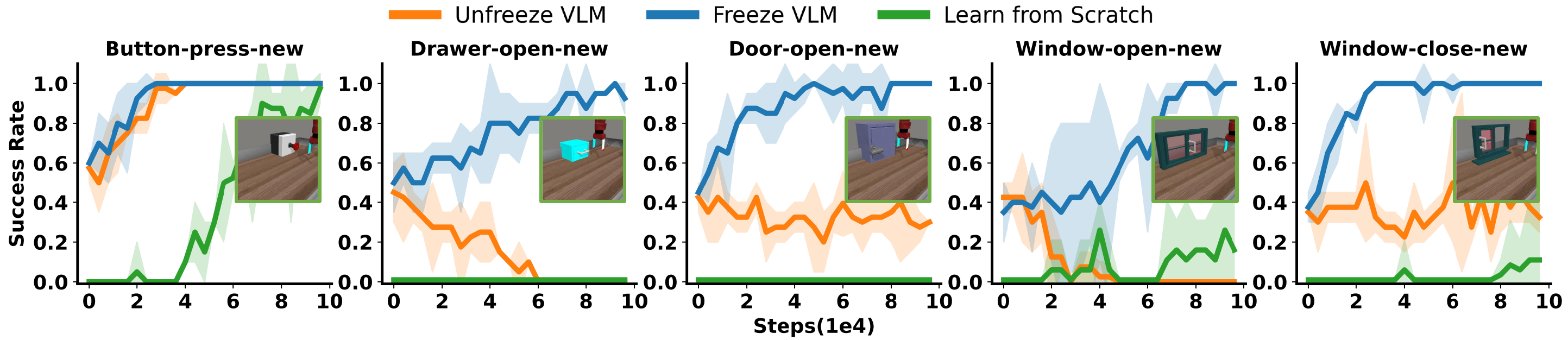}
    \vspace{-4mm}
    \caption{Reinforcement Learning process in new tasks. SFT policy can serve as a good starting point in new RL tasks compared to the learn-from-scratch policy. We also observed that fully fine-tuning VLA models can lead to performance degradation (orange lines) while freezing the VLM part can avoid collapses.}
    \label{freeze}
    \vspace{-3mm}
\end{figure*}

\subsection{Why do we adopt two-stage iterative optimization?} \label{exp2}
\textbf{Stabilizing Training Process. } 
We observed that directly fine-tuning the large VLA model using standard reinforcement learning (RL) algorithms can be unstable and lead to performance drops. As shown in Figure \ref{exp}, we observe performance drops in four out of five tasks with sparse reward in the Metaworld benchmark. This phenomenon was also observed in previous research \cite{parisotto2020stabilizing}, which encountered similar instability issues with transformer-based RL policies and had to modify transformer blocks to prevent collapse. However, these modifications are not compatible with pre-trained VLMs, instead, we freeze the VLM during the RL stage to prevent collapse.

\textbf{Managing the Model Training Burden. } 
Fully fine-tuning the VLA model with billions of parameters exceeds the computational capability of most local machines, while complete deployment on a remote server introduces parameter transmission issues and reduces the control frequency. 
Our two-stage iRe-VLA framework addresses these challenges by distributing the computational load. In the first RL stage, iRe-VLA freezes the upper-layer VLM and only adapts the lightweight action head, thus keeping computational demands affordable on the local machine. The second stage of optimization is then delegated to remote services that can handle larger computational loads. For instance, in our real-world experiments (see Section \ref{real_panda}), we conducted the RL process locally using a single NVIDIA 4090 card and performed the second stage on remote servers equipped with 4 NVIDIA A100 cards.

\begin{table*}[]
\centering
\resizebox{0.9\linewidth}{!}{
\begin{tabular}{cccccccc}
\toprule
\textbf{Metaworld} & \cellcolor{myblue}\textbf{\begin{tabular}[c]{@{}c@{}}Original \\ 25 tasks\end{tabular}} & \cellcolor{mygreen} \textbf{\begin{tabular}[c]{@{}c@{}}Button-\\ Press-new\end{tabular}} & \cellcolor{mygreen}\textbf{\begin{tabular}[c]{@{}c@{}}Drawer-\\ Open-new\end{tabular}} & \cellcolor{mygreen}\textbf{\begin{tabular}[c]{@{}c@{}}Door-\\ Open-new\end{tabular}} & \cellcolor{mygreen}\textbf{\begin{tabular}[c]{@{}c@{}}Window-\\ Open-new\end{tabular}} & \cellcolor{mygreen}\textbf{\begin{tabular}[c]{@{}c@{}}Window-\\ Close-new\end{tabular}} & \cellcolor{myred}\textbf{\begin{tabular}[c]{@{}c@{}}Unseen\\ 10 tasks\end{tabular}} \\ \hline
\textbf{SFT Policy} & \textbf{0.83} & 0.56 & 0.48 & 0.40 & 0.32 & 0.28 & 0.51 \\ \hline
\textbf{PPO-Replay} & 0.69 & 0.80 & 0.24 & 0.32 & 0.04 & 0.36 & 0.39\\ 
\textbf{iRe-VLA(Ours)} & \textbf{0.83} & \textbf{1.00} & \textbf{0.84} & \textbf{0.84} & \textbf{0.80} & \textbf{0.96} & \textbf{0.80} \\ 
\bottomrule
\end{tabular}
}
\vspace{3mm}

\centering
\resizebox{0.9\linewidth}{!}{
\begin{tabular}{cccccccc}
\toprule
\textbf{\begin{tabular}[c]{@{}c@{}}Franka \\ Kitchen\end{tabular}} & \cellcolor{myblue}\textbf{Knob-on} &\cellcolor{myblue} \textbf{Light-on} &\cellcolor{myblue} \textbf{\begin{tabular}[c]{@{}c@{}}Microwave\\ -open\end{tabular}} & \cellcolor{myblue}\textbf{\begin{tabular}[c]{@{}c@{}}Slide-door\\ -open\end{tabular}} & \textbf{\begin{tabular}[c]{@{}c@{}}\cellcolor{myblue}Left-door\\ \cellcolor{mygreen}-open\end{tabular}} & \cellcolor{mygreen}\textbf{\begin{tabular}[c]{@{}c@{}}Slide-door\\ -open-red\end{tabular}} & \cellcolor{myred}\textbf{\begin{tabular}[c]{@{}c@{}}Slide-door\\ -open-yellow\end{tabular}} \\ \hline
\textbf{SFT Policy} & 0.84 & \textbf{0.96} & 0.70 & 0.86 & 0.43 & 0.46 & \textbf{0.98} \\ \hline
\textbf{PPO-Replay} & 0.48 & 0.64 & 0.35 & 0.96 & 0.12 & 0.30 & 0.64 \\ 
\textbf{iRe-VLA(Ours)} & \textbf{0.90} & \textbf{0.98} & \textbf{0.82} & \textbf{0.99} & \textbf{0.83} & \textbf{0.99} & \textbf{1.00} \\
\bottomrule
\end{tabular}
}
\vspace{2mm}
\caption{Success rates on Metaworld and Franka-kitchen benchmark with three categories of tasks (expert tasks in \textcolor{blue}{blue}, RL-trained tasks in \textcolor{green}{green}, and unseen tasks in \textcolor{red}{red}). Standard online RL algorithms result in performance even worse than SFT policy, while iRe-VLA improves performance in three categories of tasks. } \label{table_meta}
\vspace{-4mm}
\end{table*}

\subsection{Simulated Manipulation Experiments} 
We initially conducted experiments in simulated Metaworld and Franka Kitchen benchmark, where the VLA model is first supervised on 25 tasks and 5 tasks respectively. 
VLA model can provide an effective starting point for RL tasks, accelerating the RL process compared to the learn-from-scratch approach, as demonstrated in Figure \ref{freeze}. 
Subsequently, we perform the iRe-VLA method to learn RL tasks one by one, which continuously improves the VLA model. 
We compare our method with the standard PPO algorithm \cite{schulman2017proximal}. 
To ensure a fair comparison, we also performed PPO task by task and adopted the same expert data replay strategy after each task, namely PPO-Replay. 


\textbf{Analysis. }The results are presented in Table \ref{table_meta}. Standard PPO algorithms often exhibit instability when introduced to RL tasks, as depicted in Figure \ref{freeze}. This instability not only affects performance in RL tasks but also degrades performance in previously learned tasks, even with experience replay. This decline is likely due to noisy RL gradients that adversely affect the pre-trained representations within the VLA model. In contrast, our two-stage iRe-VLA method stabilizes the RL process and effectively enhances task performance across both seen and unseen tasks. The advantage of the iRe-VLA method can be reflected in three aspects:

\textbf{(1) Improved Performance in Original Tasks. }We can continue to improve performance in seen expert tasks through online interaction. For instance, in the Franka-kitchen benchmark, the supervised VLA model achieved a modest success rate in the expert task \textit{left-door-open} due to limited demonstrations. Our iRe-VLA method improves the success rate of this task from 0.43 to 0.83.

\textbf{(2) Improved Performance in RL Tasks. }It is crucial for intelligent agents to adapt to tasks excluded in expert data autonomously. We explored various RL tasks (as detailed in the second column of Figure \ref{exp}) and applied our iterative RL algorithm to address these tasks. As indicated in Table \ref{exp}, our iRe-VLA method successfully tackled new tasks in each domain without catastrophic forgetting \cite{mccloskey1989catastrophic}.

\textbf{(3) Improved Generalization in Unseen Tasks. }In addition to the enhanced performance in RL-trained tasks 
through online iterations, we also observed increased success rates in unseen tasks, indicating better generalization ability. As the agent tackles an increasing variety of tasks automatically, its generalization ability correspondingly strengthens. For example, after mastering four types of window tasks in Metaworld, the agent effectively generalized to windows of unseen colors and shapes.

\begin{figure}[h]
    \centering
    \vspace{-2mm}
    \includegraphics[width=0.35\textwidth]{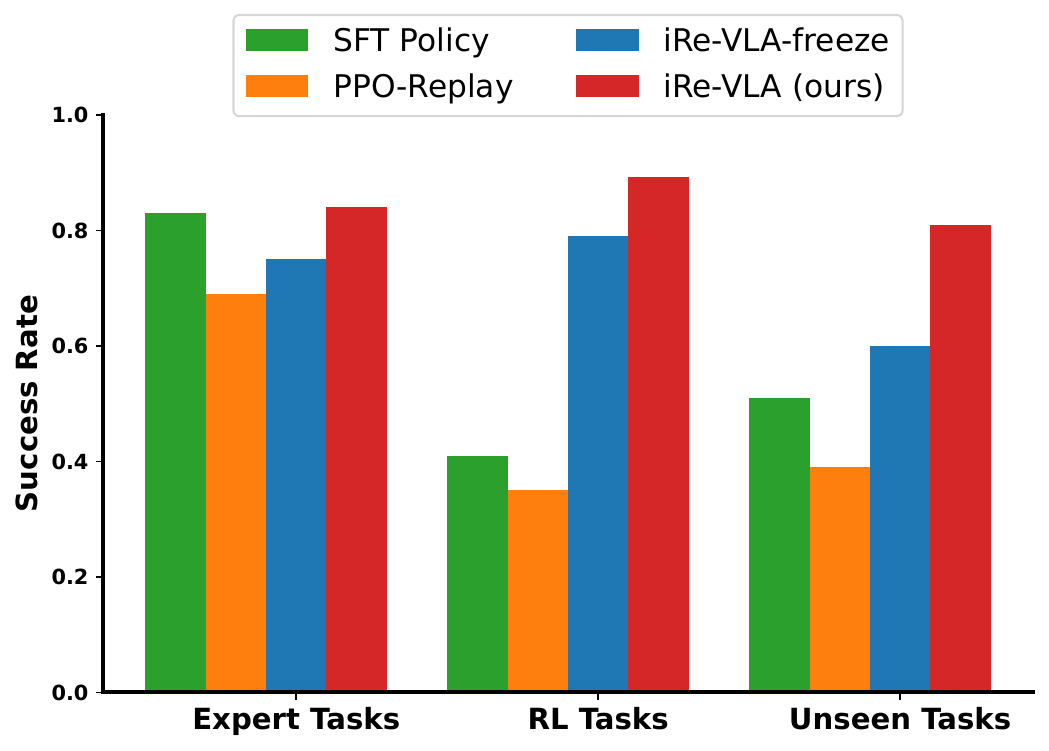}
    \vspace{-2mm}
    \caption{Ablations. Freezing VLM all the time leads to performance drops.}\label{ablation}
    \label{real_exp}
    \vspace{-5mm}
\end{figure}

\textbf{Ablation Study. }In the iRe-VLA method, the whole VLM is trainable in the second supervised learning stage. We conducted ablation studies by freezing the VLM in both stages, namely iRe-VLA-freeze. In this way, online iteration data can not affect the VLM latent. The outcomes, depicted in Figure \ref{ablation}, suggest that permanently freezing the VLM leads to a reduction in performance. This could be attributed to the action head's limited expressiveness compared to the full VLA model. Additionally, online robotic action data could enhance the representations in the upper-layer VLM, thereby augmenting the VLA model's generalizability in unseen tasks, while freezing VLM in both stages can not improve the VLM representation.

\begin{figure}[t]
    \centering
    \includegraphics[width=0.48\textwidth]{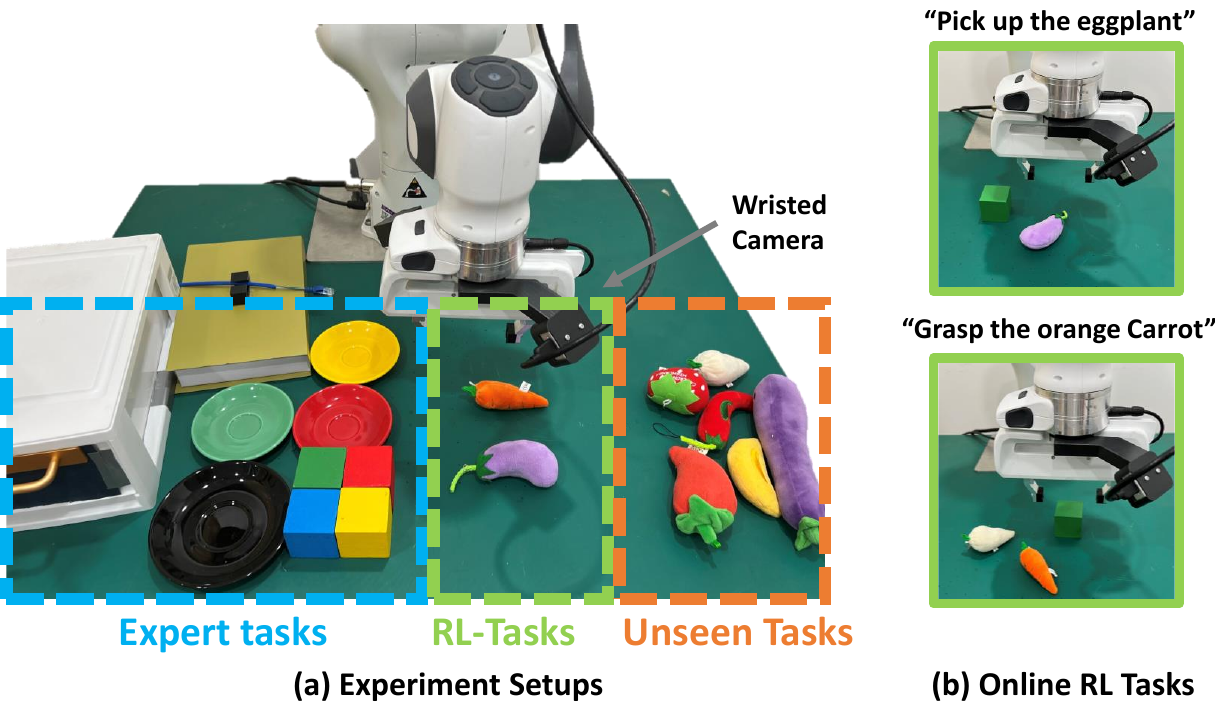}
    \includegraphics[width=0.38\textwidth]{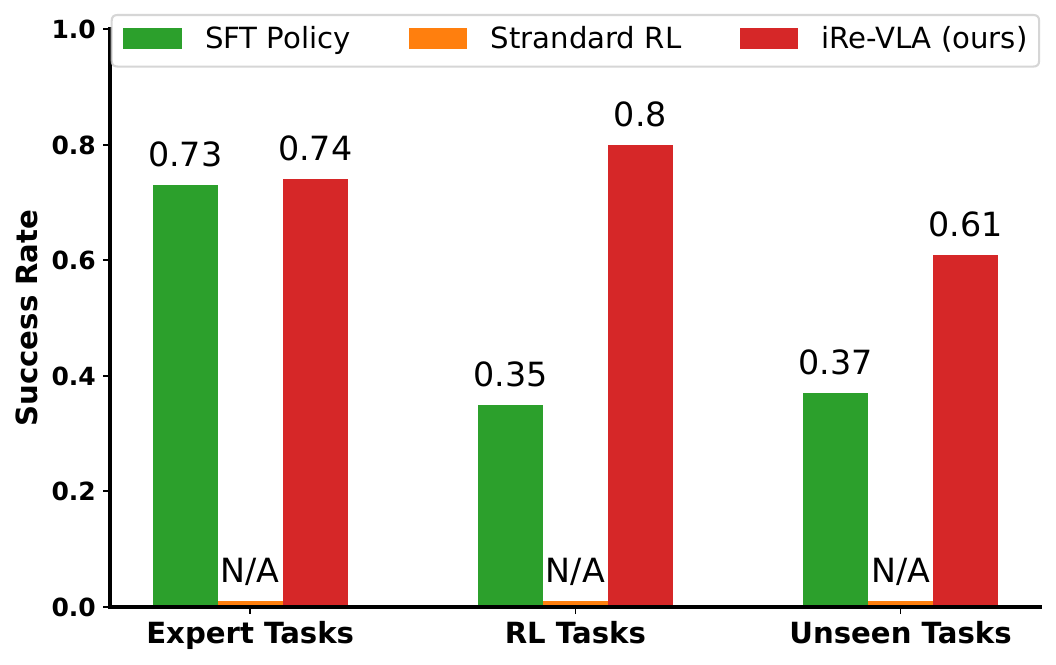}
    \vspace{-2mm}
    \caption{Real-world experiments with panda arm. We did not report the standard RL results in real-world tasks since directly fine-tuning the entire VLA model exceeded the computational capabilities of our local machine.}
    \label{real_exp}
    \vspace{-4mm}
\end{figure}

\subsection{Real-world Manipulation Experiments}\label{real_panda}
\textbf{Experiment Setups. }Our real-world experiment follows the set ups described in SERL \cite{luo2024serl,luo2024fmb}, a useful software suite for the real-world RL.  
We first train a VLA model on 2,000 human-collected expert data across various task categories, including pick (grasp), place, button-press, cable-route, and drawer operations. 

We notice that the learned VLA model shows a certainty success rate on unseen objects thanks to the generalization ability of the VLA model. Then we adopt online RL to further increase the success rate on unseen objects. 
We implemented several key design choices to enhance sample efficiency and ensure computational affordability within the context of large Vision-Language-Action (VLA) models. To improve sample efficiency, we adopted the SACfD algorithm \cite{vecerik2017leveraging,haarnoja2018soft}. Specifically, when introduced to a new task, we initially utilize zero-shot transferred VLA models to collect a demonstration buffer containing 20 successful trajectories. During training, we sample 50\% transitions from the demonstration buffer and 50\% from the online buffer, as outlined in \cite{luo2024serl}. To manage computational costs, each image observation is processed by the VLM only once, and the resulting latent output is stored in the buffer. Subsequently, we implement the SACfD algorithm in this latent space.

\textbf{Results. }The expert pick demonstrations were limited to blocks of four colors, and we extended the online RL to objects with irregular shapes, such as eggplants and carrots. The real-world RL training process for each new task costs around one hour, similar to time costs in SERL \cite{luo2024serl}. The success rates before and after RL process are shown in Figure \ref{real_exp}, our iRe-VLA pipeline increased the success rate for picking eggplants or carrots from 0.35 to 0.80. Moreover, the success rates for the original tasks remained stable, and the picking success rate for unseen objects also improved from 0.37 to 0.61.

\section{Conclusion and Limitation}
In this paper, we explore ways to further enhance the VLA model through online reinforcement learning. Fine-tuning large VLA models presents several challenges, but our proposed iRe-VLA methods stabilize the training process and significantly reduce computational demands. Experiments on both simulated and real-world manipulation tasks confirm the effectiveness of iRe-VLA. A potential limitation is that it can only improve skills within seen types and cannot learn entirely new skills under sparse-reward online RL conditions.

\bibliographystyle{IEEEtran}
\bibliography{IEEEexample}

\addtolength{\textheight}{-7cm}   

\end{document}